\documentclass[conference]{IEEEtran}
\usepackage{newtxtext,newtxmath} 
\usepackage[numbers]{natbib}
\usepackage{multicol}
\usepackage[bookmarks=true]{hyperref}
\usepackage{booktabs,multirow}
\usepackage{siunitx}
\usepackage{graphicx}
\usepackage{flushend}
\usepackage{subcaption}
\usepackage{stfloats}
\begin{document}
\title{Tiered-Resolution Volumetric Mapping for Real-Time Simultaneous Planning and Mapping on Natural Outdoor Environments}
\author{
    \IEEEauthorblockN{
        Afonso E. Carvalho,
        David Portugal,
        Paulo Peixoto
    }
    \IEEEauthorblockA{
        \textit{Institute of Systems and Robotics}\\
        \textit{University of Coimbra}\\
        Coimbra, Portugal
    }
    \and
    \IEEEauthorblockN{George Kantor}
    \IEEEauthorblockA{
        \textit{Robotics Institute}\\
        \textit{Carnegie Mellon University}\\
        Pittsburgh, PA, USA
    }
}
\maketitle
\begin{abstract}
Real-time volumetric mapping for a planning robot is dominated by the cost of integrating and storing information at the finest resolution, yet only a small fraction of any map ever lies on or near a path the planner actually considers.
We exploit this with a tiered-resolution mapping mechanism that integrates LiDAR into a single map structure at two resolutions simultaneously: fine inside a corridor around the robot's current plan, and coarse everywhere else.
As opposed to traditional navigation pipelines where path planning uses the mapping result and provides no feedback, we propose a feedback loop where planning also informs mapping to dynamically integrate information at different leaf resolutions.
Across two real-world teleoperated datasets, replayed and scored offline against an independent full-resolution map, the tiered maps are \textbf{13--15$\times$ smaller} and require \textbf{6--8$\times$ less integration CPU time} than a uniform fine map, while producing paths at a similar cost per meter as the ones generated from fine resolution maps on both scenarios.
A uniform coarse map reaches the same size savings but its path quality collapses (\textbf{+210\% cost per meter}) on the harder scenario, because it over-discretizes terrain the robot must traverse.
Tiering approximates the coarse map's cheapness without its failure mode.
\end{abstract}
\IEEEpeerreviewmaketitle

\begin{figure*}[b]
\centering
\begin{subfigure}[t]{0.48\linewidth}
\centering
\includegraphics[width=\linewidth]{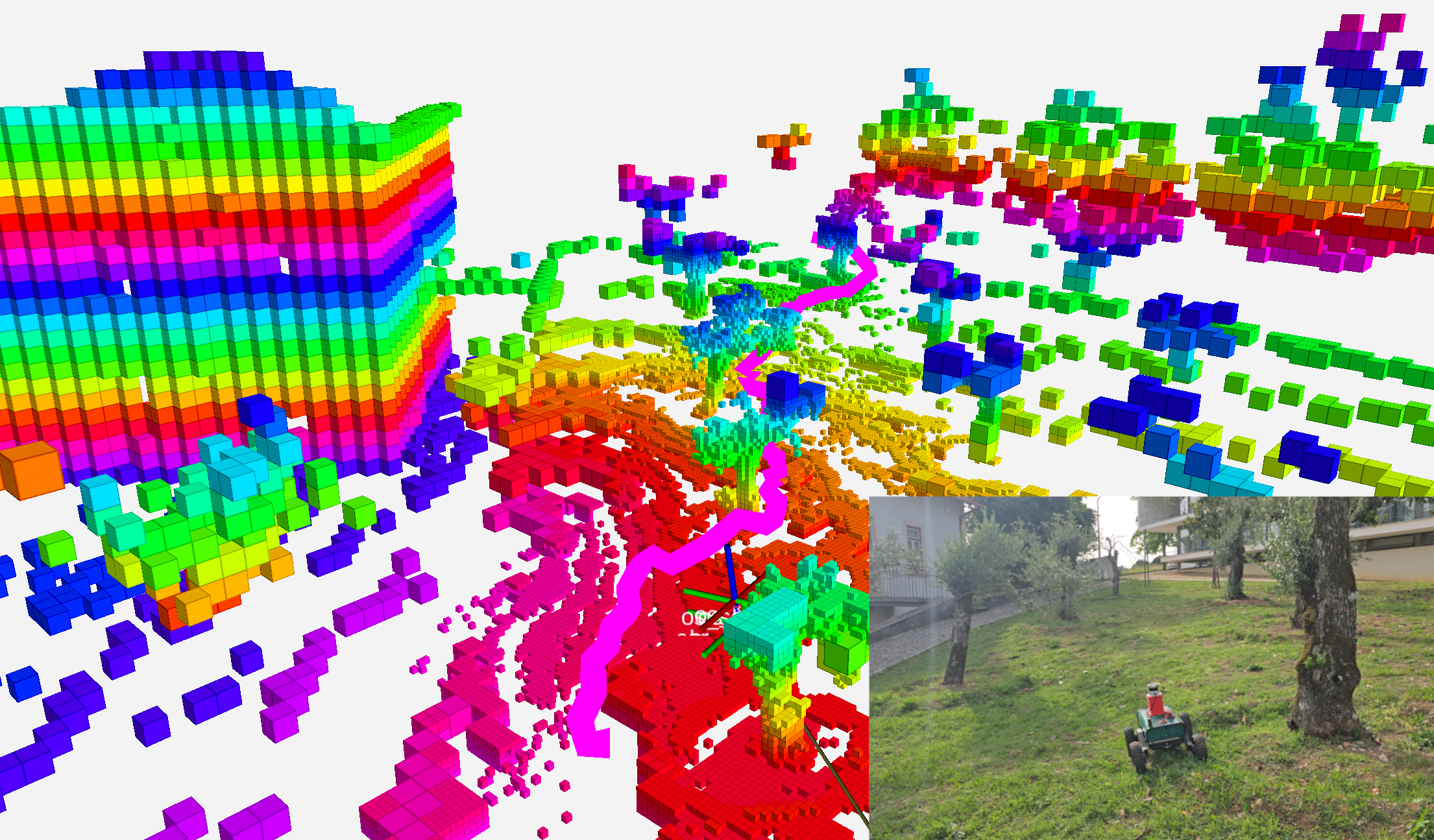}
\caption{Partial map from the first experimental scenario.}
\label{fig:scenario1_tiered}
\end{subfigure}\hfill
\begin{subfigure}[t]{0.48\linewidth}
\centering
\includegraphics[width=\linewidth]{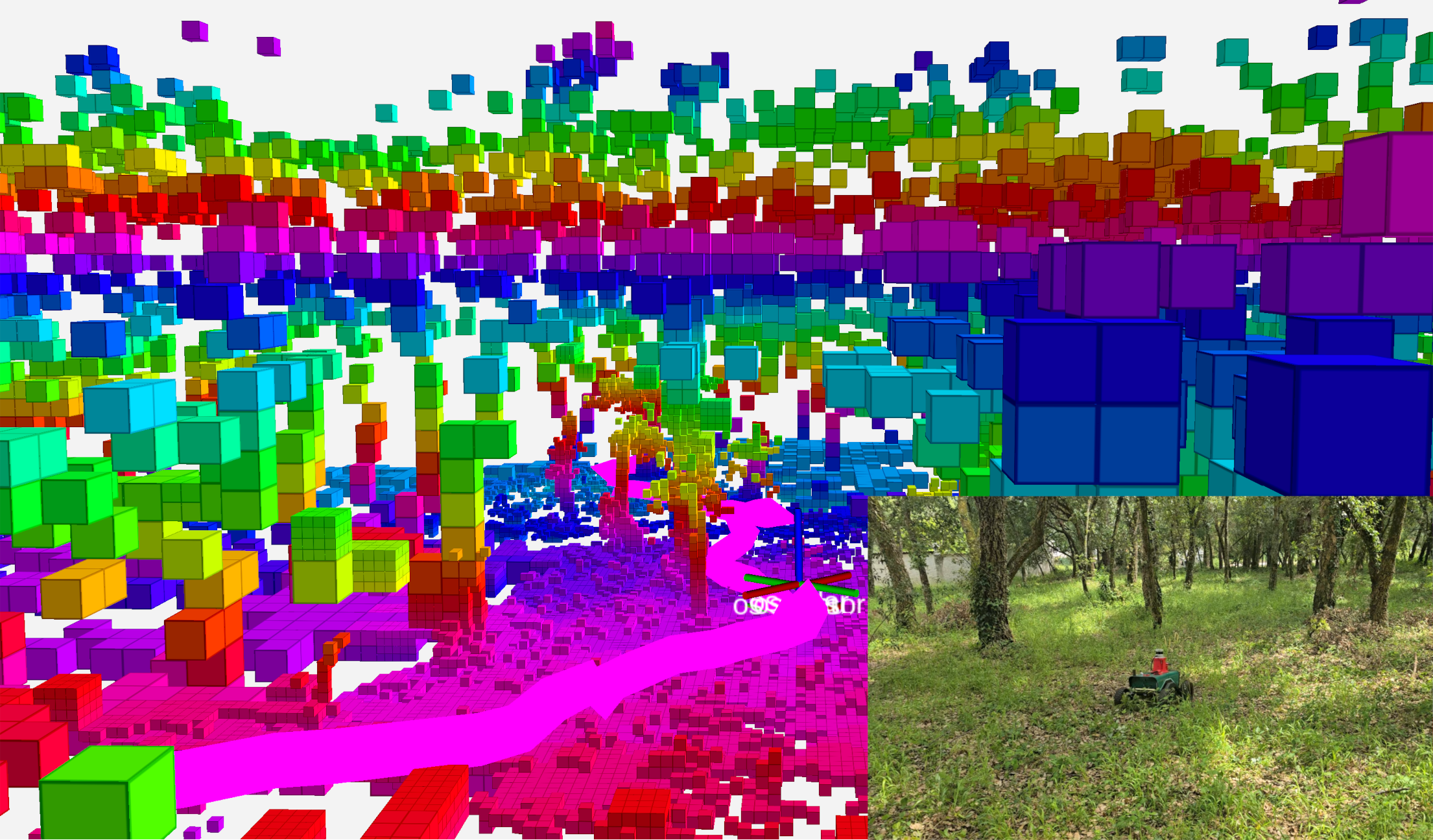}
\caption{Partial map from the second experimental scenario.}
\label{fig:scenario2_tiered}
\end{subfigure}
\caption{Part of the volumetric map generated with fine resolution near the path (3m corridor), and coarse resolution otherwise. The path is colored magenta, and the voxels are colored by height.}
\label{fig:scenarios_tiered}
\end{figure*}

\section{Introduction}

Volumetric occupancy mapping is a core building block of autonomous navigation, and its real-time cost is dominated by the finest resolution at which measurements are integrated and stored~\cite{hornung2013octomap,oleynikova2017voxblox,reijgwart2023wavemap}.
A system that uses centimeter-scale fidelity everywhere also incurs it's cost everywhere, even though most of a large environment is never on, or even close to, the area the robot will actually navigate on.

\begin{figure*}[t]
\centering
\includegraphics[width=\linewidth]{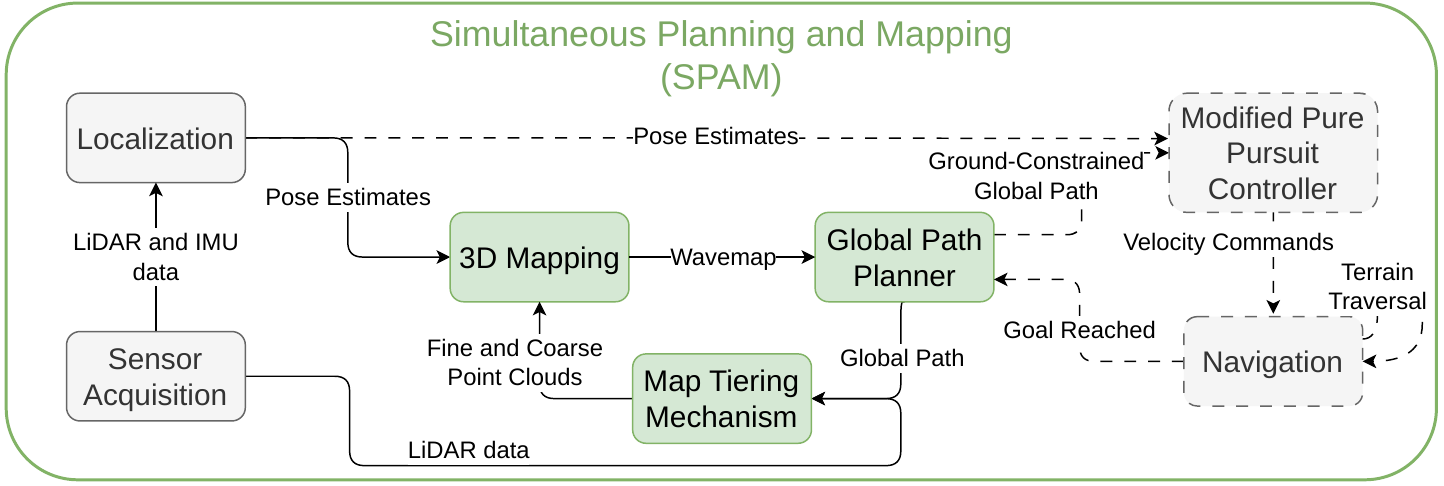}
\caption{The proposed architecture. Dashed lines represent the modules that would be added to perform actuation, and green modules represent the feedback loop described.}
\label{fig:architecture}
\end{figure*}

This paper turns that observation into a mechanism: allocate resolution by plan-relevance, finely mapping near the plan and coarsely everywhere else, as shown in Fig.~\ref{fig:scenarios_tiered}, according to the diagram presented in Fig.~\ref{fig:architecture}.
The natural objection -- why not map everything coarse? -- is exactly the discriminator our experiments are built around: a uniform coarse map quantizes the terrain the robot must traverse, degrading and over-optimizing the paths the planner reasons about.
Tiering is only worth defending if it avoids that failure while approaching the coarse map's savings, and on the harder of our two real-world scenarios, that is precisely the trade it achieves.

This paper contributes:

\begin{itemize}
    \item An online, plan-conditioned resolution-allocation mechanism and a resolution router that split each LiDAR sweep into a fine and a coarse stream by distance from the robot center or the live planned-path corridor and integrates both into a single hashed-wavelet octree via two simultaneous integrators;
    \item An evaluation methodology isolating resolution allocation as the sole independent variable across six configurations, scored offline under a shared, independent reference map;
    \item A secondary ablation against a plan-free, budget-matched sensor-range baseline.
\end{itemize}

Section~\ref{sec:related} situates this against prior mapping work; Section~\ref{sec:method} describes the mechanism; Section~\ref{sec:experiments} presents the evaluation's results and discussion, and Section~\ref{sec:conclusion} presents the conclusion and future work.

\section{Background and Related Work}
\label{sec:related}
Octree-based mapping frameworks such as OctoMap~\cite{hornung2013octomap}, Voxblox~\cite{oleynikova2017voxblox}, and UFOMap~\cite{duberg2020ufomap} allocate resolution uniformly or as a fixed function of the data structure, rather than in response to the task. Wavemap~\cite{reijgwart2023wavemap}, which our system uses as the mapping framework, generalizes this through wavelet compression on a hashed-wavelet octree supporting multiple simultaneous resolutions in one tree and providing access to multiple integrators, which have been used in the original publication to integrate information from different sensors, such as depth cameras and LiDARs.
Closest to our work, Funk~et~al.~\cite{funk2021multires} choose resolution per frame via max-min pooling of the depth image, coupling resolution to local sensor geometry rather than the planner's current intent, and adaptive-resolution mapping has been explored for energy-constrained aerial navigation along regions of high traversal probability~\cite{liu2025eans}; neither conditions the split on a planner's live output nor scores path quality against an independent oracle.
Bandwidth-constrained map compression for multi-robot communication~\cite{Psomiadis2023} shares our principle -- not all of a map is equally worth preserving at full fidelity -- for a different resource.
We plan in a continuous traversability field with sampling-based, asymptotically optimal planners from the RRT*/BIT* family~\cite{karaman2011rrtstar,gammell2015bitstar}, as also demonstrated for large-scale outdoor ground-robot navigation using Wavemap-based volumetric mapping~\cite{Carvalho2026spam} -- this choice is orthogonal to our contribution, which only needs the planner to expose its current path.

\section{Methodology}
\label{sec:method}

\begin{figure*}[t]
\centering
\begin{subfigure}[t]{0.485\linewidth}
\includegraphics[width=\linewidth]{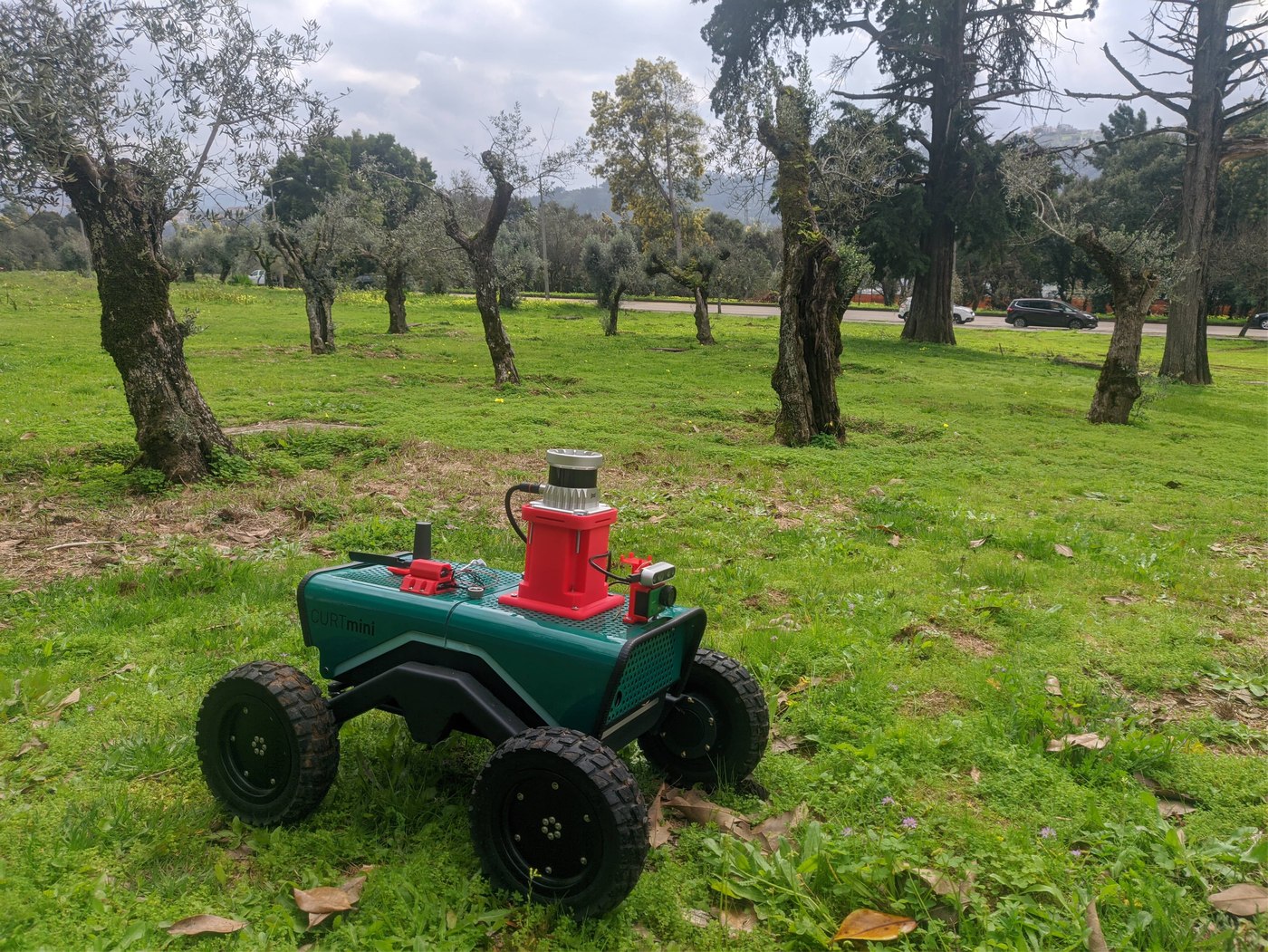}
\caption{Scenario 1, a sparse olive orchard with slightly sloped, grassed terrain.}
\end{subfigure}\hfill
\begin{subfigure}[t]{0.485\linewidth}
\includegraphics[width=\linewidth]{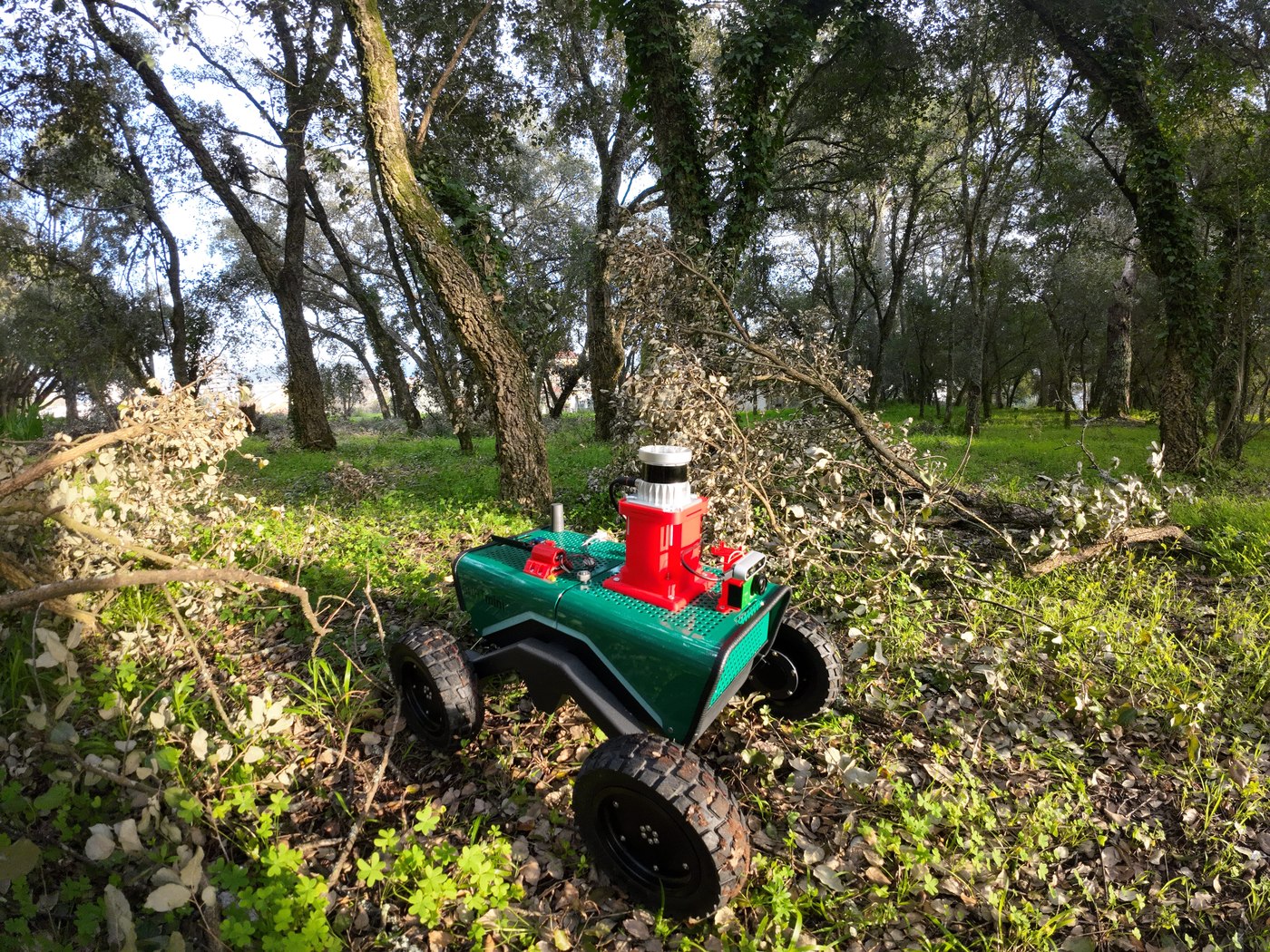}
\caption{Scenario 2, a denser wooded terrain with downed trees and tree branches.}
\end{subfigure}
\caption{The CURTmini platform in the two experimental environments.}
\label{fig:robots}
\end{figure*}

The system under test, SPAM (Simultaneous Planning And Mapping), presented in Fig.~\ref{fig:architecture}, runs a volumetric mapper and a sampling-based planner concurrently in ROS.
Mapping is performed with a custom fork of Wavemap: a single hashed-wavelet octree with a base minimum voxel width of~\SI{0.10}{\meter}, written by one or more integrators, each capped at a configurable maximum update resolution that it can never refine past.
Planning uses a hybrid approach of RRT* and BIT*.
RRT* is used to generate paths when the final goal is not in known terrain, returning the best possible approximate path in the goal's direction; BIT* is used when the terrain where the goal sits becomes mapped, to generate an exact solution, considering that it converges faster to a better solution than RRT*.

The map tiering mechanism is a ROS node that subscribes to the raw LiDAR cloud and the planner's current path and, on every sweep: (1) computes each point's distance from the corridor swept out by the current plan; (2) splits the sweep into a fine cloud within a configurable high-resolution distance (hrd) of the corridor and a coarse cloud for everything else; (3) integrates both into the same octree via two simultaneous integrators, capped at \SI{0.10}{\meter} and \SI{0.40}{\meter}, respectively. 
We call this \texttt{split\_mode: path}, the system under test; a \texttt{split\_mode: sensor\_range} variant replaces distance-from-plan with distance-from-sensor as a plan-free baseline.
Resolution allocation is the only variable that differs across configurations: every arm shares the same \SI{0.10}{\meter} tree, Ouster projection model ($128\times1024$), \texttt{continuous\_beam} measurement model, and planner, so any measured difference is attributable to how resolution was allocated.

The planner and offline scorer share one 2.5D traversability objective, cost per unit length $c = 1 \cdot c_{\text{length}} + 4 \cdot c_{\text{slope}} + 2 \cdot c_{\text{roughness}}$, with $z$ pinned to ground height.
Preliminary testing regarding the traversability weighted function showed that, for fixed endpoints, a pure-length route can cost up to $\sim$63 terrain-units versus $\sim$29 for the full objective on a route only $\sim$6\% longer.

\section{Experimental Validation}
\label{sec:experiments}

\subsection{Setup}
\begin{figure*}[t]
\centering
\includegraphics[width=\linewidth]{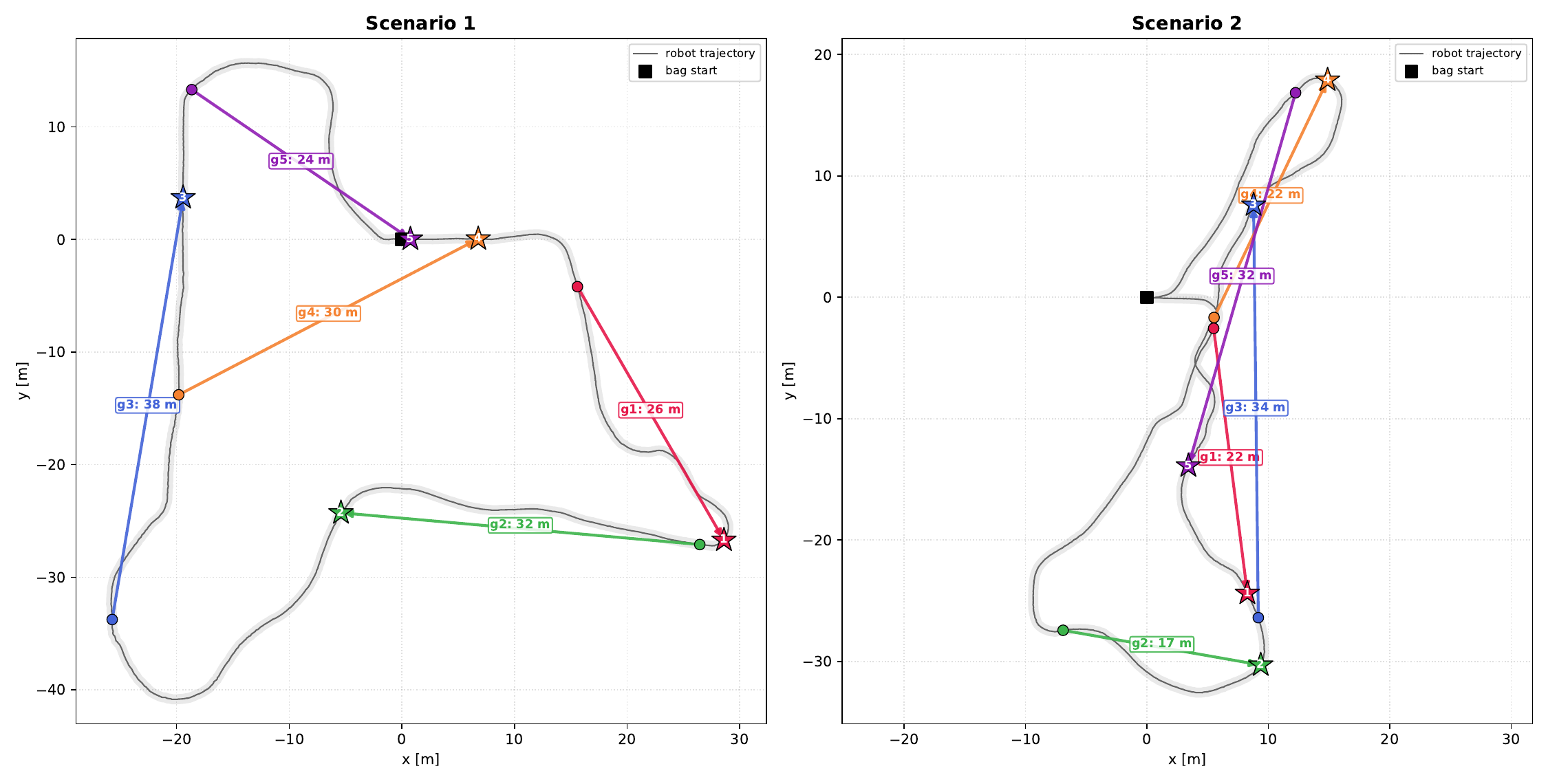}
\caption{Recorded trajectories and the five goal poses (planning probes) issued per scenario.}
\label{fig:goals}
\end{figure*}

We use two real-world LiDAR datasets both collected using an Ouster OS1-128, $128\times1024$@10 Hz, recorded by teleoperating the CURTmini ground robot, shown in Fig.~\ref{fig:robots} through natural terrain: Scenario~1 is a sparse olive-tree orchard, and Scenario~2 is a denser wooded terrain with downed trees and branches).
Scenario~2 is the harder of the two and is where routes actually cross coarsened terrain; Scenario~1's short queries cross little of it, making it a near-control.
Fig.~\ref{fig:goals} shows both trajectories with the five goal poses used per scenario.

We evaluate six experimental arms by replaying these datasets and running localization, mapping and path planning in real time, isolating resolution allocation as the only variable: \texttt{uniform\_fine} (\SI{0.10}{m} everywhere, best-quality reference) and \texttt{uniform\_coarse} (\SI{0.40}{m} everywhere, cheapest mapping); \texttt{tiered} (\texttt{hrd}~=~\SI{3.0}{m}, the system under test) plus \texttt{tiered\_hrd1p5} and \texttt{tiered\_hrd6p0} to isolate the effect of the corridor width; and a plan-free \texttt{range\_r\{6,5\}p0} baseline (fine near \emph{sensor} instead of plan), radius budget-matched to \texttt{tiered}'s node count on each scenario.
The laptop used for these experiments is an Asus ROG Strix Scar G16, equipped with an Intel Core Ultra 9 275HX processor and 64GB of RAM.

Each run replays a recorded teleoperation bag rather than driving live, so the robot's pose sequence is identical across every run and any inter-run difference is attributable to the map, not to divergent trajectories (odometry from iG-LIO).
Five goals per scenario (Fig.~\ref{fig:goals}) are issued at fixed sim-clock timestamps at poses the robot actually occupied and act as planning probes, not navigation waypoints.
Six configurations $\times$ 5 repeats $\times$ 2 scenarios gives 60 runs, all completed, zero failures.

Route quality is scored offline under a seventh, independent control map per scenario (uniform-fine, driven through with no goal ever issued, so it cannot be biased by any arm's own planning). We then obtain path's planned on each configuration's own map and re-score them under the control; re-scoring the same paths under the arm's own map, with the same objective, yields a confidence cost, giving an \emph{optimism} ratio $c_{\text{control}}/c_{\text{confidence}}$ -- how much an arm's own map flattered itself.
A same-map control gives optimism $\approx1.000$.
We report map cost (octree nodes/memory, mapping CPU, process RSS) and route quality over the matched set (cost per meter of progress, exact-solve fraction, optimism), with 95\% confidence intervals (CI) across 5 repeats against \texttt{uniform\_fine} (\textbf{ns}: CI spans zero; \textbf{sig}: excludes zero).

\subsection{Results and Discussion}
\label{sec:results}

\begin{figure*}[t]
\centering
\begin{subfigure}[t]{0.48\linewidth}
\centering
\includegraphics[width=\linewidth]{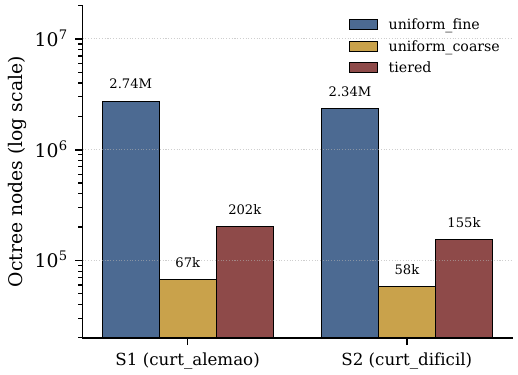}
\caption{Map size (log scale), both scenarios.}
\label{fig:mapcost}
\end{subfigure}\hfill
\begin{subfigure}[t]{0.48\linewidth}
\centering
\includegraphics[width=\linewidth]{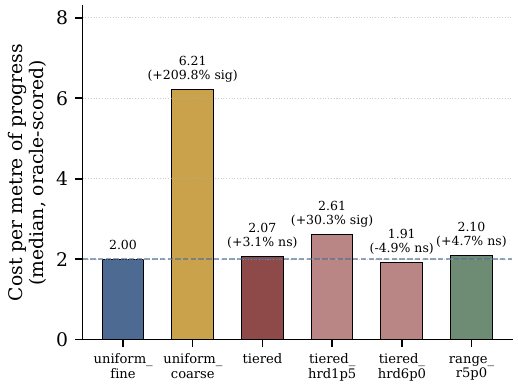}
\caption{Cost per meter of progress, Scenario 2.}
\label{fig:routequality}
\end{subfigure}
\caption{Tiered tracks close to uniform coarse in map size while matching uniform fine in path quality on the harder scenario; uniform coarse collapses.}
\label{fig:combined}
\end{figure*}

Table~\ref{tab:mapcost} and Fig.~\ref{fig:mapcost} report map cost.
Tiering cuts octree nodes by 92.7\% (13.6$\times$) on Scenario 1 and 93.4\% (15.1$\times$) on Scenario 2, with matching memory reductions. 
Mapping CPU (summed across the mapper and, for tiered, also the router, so its overhead is taken into account) drops 83.8\%/87.6\%, i.e., 6--8$\times$ less CPU usage during integration.
Process RSS drops less (38--44\%) since a fixed ROS/odometry baseline dominates it.

Table~\ref{tab:quality} and Fig.~\ref{fig:routequality} report route quality (25/25 coverage, every configuration).
On Scenario 2, the discriminator, uniform coarse costs 209.8\% more per meter and solves only 40\% of queries exactly, while tiered, at essentially the same $\sim$93\% size saving, matches fine (+3.1\%, ns) with exact-solve reasonably close (0.576 vs.\ 0.736) -- i.e., the coarse map's size close to the fine map's quality.
On Scenario 1, coarsening has little effect even for uniform coarse (+5.7\%, ns), so tiered's own +20.6\% (also \textit{ns}) point estimate, whose CI spans zero at $n=5$ is not considered a regression, meaning that tiering is most likely to help when a route must actually traverse coarse terrain, which Scenario 1 mostly avoids and Scenario 2 forces.
Optimism corroborates this, since uniform coarse flatters itself (1.15 to 1.21, hiding cost it would actually pay), while tiered stays near 1.0 (1.013 on Scenario 2).

\begin{table}[t]
\centering
\caption{Map cost: tiered vs.\ uniform maps. All differences vs.\ \texttt{uniform\_fine} have 95\% CIs excluding zero.}
\label{tab:mapcost}
\footnotesize
\setlength{\tabcolsep}{4.5pt}
\begin{tabular}{llrrrr}
\toprule
Metric & Scen. & Fine & Coarse & Tiered & vs.\ fine \\
\midrule
Nodes            & S1 & 2{,}744{,}307 & 67{,}351 & 201{,}632 & $-$92.7\% \\
Nodes            & S2 & 2{,}340{,}180 & 57{,}827 & 154{,}834 & $-$93.4\% \\
Memory (MB)      & S1 & 135.7 & 3.60 & 10.29 & $-$92.4\% \\
Memory (MB)      & S2 & 116.6 & 3.01 & 7.86  & $-$93.3\% \\
Map CPU (\%)      & S1 & 219.3 & 22.9 & 35.4  & $-$83.8\% \\
Map CPU (\%)      & S2 & 288.0 & 20.1 & 35.8  & $-$87.6\% \\
Process RSS (MB)  & S1 & 432.7 & 249.0 & 244.5 & $-$43.5\% \\
Process RSS (MB)  & S2 & 373.5 & 217.0 & 231.9 & $-$37.9\% \\
\bottomrule
\end{tabular}
\end{table}

\begin{table}[t]
\centering
\caption{Route quality (25/25 coverage, every arm). The "vs.\ fine" column is the paired 95\% CI difference.}
\label{tab:quality}
\footnotesize
\setlength{\tabcolsep}{4.5pt}
\begin{tabular}{llrrrr}
\toprule
Arm & Scen. & Cost/m & vs.\ fine & Exact & Optim. \\
\midrule
uniform\_fine     & S1 & 3.39 & --                 & 0.632 & 1.009 \\
uniform\_coarse   & S1 & 3.59 & +5.7\% ns          & 0.176 & 1.211 \\
\textbf{tiered}   & S1 & 4.09 & +20.6\% ns         & 0.568 & 1.211 \\
tiered\_hrd1p5    & S1 & 4.10 & +20.8\% ns         & 0.536 & 1.252 \\
tiered\_hrd6p0    & S1 & 4.07 & +19.9\% ns         & 0.544 & 1.143 \\
range\_r6p0       & S1 & 3.78 & +11.3\% ns         & 0.352 & 1.228 \\
\midrule
uniform\_fine     & S2 & 2.00 & --                 & 0.736 & 1.022 \\
uniform\_coarse   & S2 & 6.21 & +209.8\% \textbf{sig} & 0.400 & 1.150 \\
\textbf{tiered}   & S2 & 2.07 & +3.1\% ns          & 0.576 & 1.013 \\
tiered\_hrd1p5    & S2 & 2.61 & +30.3\% \textbf{sig}  & 0.640 & 0.994 \\
tiered\_hrd6p0    & S2 & 1.91 & $-$4.9\% ns        & 0.808 & 1.045 \\
range\_r5p0       & S2 & 2.10 & +4.7\% ns          & 0.864 & 1.051 \\
\bottomrule
\end{tabular}
\end{table}

Sweeping \texttt{hrd} shows the trade is genuine, as an \texttt{hrd} of 1.5~m gives a smaller map (149{,}977 vs.\ 201{,}632 nodes on Scenario 1) but a significant Scenario 2 regression (+30.3\%), while 6.0~m gives a larger map (299{,}215 nodes) and the closest-to-fine Scenario 2 quality ($-$4.9\%, ns); \SI{3.0}{m} sits at a sensible middle, $\sim$93\% saving at fine parity. Budget-matched to tiered's node count, the plan-free range baseline is competitive on cost on both datasets (Scenario 1 +11.3\%, Scenario 2 +4.7\%, both ns) and exceeds tiered on Scenario 2 exact-solve (0.864 vs.\ 0.576), though tiered leads on Scenario 1 exact-solve (0.568 vs.\ 0.352).
These results support the claim of the advantage of tiered approaches over uniform strategies, with the range comparison a secondary ablation on whether plan feedback specifically is doing the work in these scenarios.
In this case, it is not performing significantly different, but we expect this to change when testing in a closed-loop setup with longer paths, which we mention as future work.

\section{Conclusion and Future Work}
\label{sec:conclusion}
Tiered-resolution mapping exploits the assumption that, especially when using asymptotically optimal path planners, the area surrounding the proposed plan at each iteration is likely to contain a lower cost path, should one exist.
Furthermore, refining the area around the proposed plan can also help to refine it in the next iterations against real terrain topography that was filtered away by the initial coarse resolution.
Integrating LiDAR fine inside the planned-path corridor and coarse elsewhere achieves 13--15$\times$ smaller maps and 6--8$\times$ less integration CPU than uniform fine, at a similar path quality to that achieved using fine quality on both real-world scenarios, while a uniform coarse map of the same size collapses (+210\%) on the scenario whose routes actually cross coarsened terrain, actively misleading the planner about route cost by our optimism metric.
Priority next steps are closed-loop validation, even more challenging outdoor scenarios and a higher amount of experimental runs to tighten Scenario 1's confidence interval, and explore a hybrid plan/sensor-range split to capture immediate terrain topography as well as farther-away features in the vicinity of the planned path.

\section*{Acknowledgments}
This work was supported by the Fundação para a Ciência e a Tecnologia (FCT) under the Affiliated Ph.D. CMU Portugal Program, ref. PRT/BD/153920/2022. Support for this research was also provided by FCT through the ISR's Multiannual Funding for R\&D Units (UIDB/00048/2025, DOI: \url{https://doi.org/10.54499/UID/00048/2025}), and by the RUGGED Project (2023.16314.ICDT, DOI: \url{https://doi.org/10.54499/2023.16314.ICDT}).

\linespread{1.9}
\bibliographystyle{plainnat}
\bibliography{references}

@inproceedings{reijgwart2023wavemap,
  author    = {Reijgwart, Victor and Cadena, Cesar and Siegwart, Roland and Ott, Lionel},
  title     = {Efficient Volumetric Mapping of Multi-Scale Environments Using Wavelet-Based Compression},
  booktitle = {Proceedings of Robotics: Science and Systems},
  year      = {2023},
  address   = {Daegu, Republic of Korea},
  month     = {July}
}

@article{hornung2013octomap,
  author  = {Hornung, Armin and Wurm, Kai M. and Bennewitz, Maren and Stachniss, Cyrill and Burgard, Wolfram},
  title   = {{OctoMap}: An Efficient Probabilistic {3D} Mapping Framework Based on Octrees},
  journal = {Autonomous Robots},
  year    = {2013},
  volume  = {34},
  number  = {3},
  pages   = {189--206}
}

@inproceedings{oleynikova2017voxblox,
  author    = {Oleynikova, Helen and Taylor, Zachary and Fehr, Marius and Siegwart, Roland and Nieto, Juan},
  title     = {Voxblox: Incremental {3D} {E}uclidean Signed Distance Fields for On-Board {MAV} Planning},
  booktitle = {IEEE/RSJ International Conference on Intelligent Robots and Systems (IROS)},
  year      = {2017},
  pages     = {1366--1373}
}

@article{duberg2020ufomap,
  author  = {Duberg, Daniel and Jensfelt, Patric},
  title   = {{UFOMap}: An Efficient Probabilistic {3D} Mapping Framework That Embraces the Unknown},
  journal = {IEEE Robotics and Automation Letters},
  year    = {2020},
  volume  = {5},
  number  = {4},
  pages   = {6411--6418}
}

@article{funk2021multires,
  author  = {Funk, Nils and Tarrio, Juan and Papatheodorou, Sotiris and Popovi\'{c}, Marija and Alcantarilla, Pablo F. and Leutenegger, Stefan},
  title   = {Multi-Resolution {3D} Mapping With Explicit Free Space Representation for Fast and Accurate Mobile Robot Motion Planning},
  journal = {IEEE Robotics and Automation Letters},
  year    = {2021},
  volume  = {6},
  number  = {2},
  pages   = {3553--3560}
}

@article{liu2025eans,
  author  = {Liu, Tian and Liu, Han and Li, Boyang and Chen, Long and Huang, Kai},
  title   = {{EANS}: Reducing Energy Consumption for {UAV} with an Environmental Adaptive Navigation Strategy},
  journal = {arXiv preprint arXiv:2506.20485},
  year    = {2025}
}

@article{Psomiadis2023,
  author  = {Psomiadis, Evangelos and Maity, Dipankar and Tsiotras, Panagiotis},
  title   = {Communication-Aware Map Compression for Online Path-Planning},
  journal = {arXiv preprint arXiv:2309.13451},
  year    = {2023}
}

@article{karaman2011rrtstar,
  author  = {Karaman, Sertac and Frazzoli, Emilio},
  title   = {Sampling-Based Algorithms for Optimal Motion Planning},
  journal = {The International Journal of Robotics Research},
  year    = {2011},
  volume  = {30},
  number  = {7},
  pages   = {846--894}
}

@inproceedings{gammell2015bitstar,
  author    = {Gammell, Jonathan D. and Srinivasa, Siddhartha S. and Barfoot, Timothy D.},
  title     = {Batch Informed Trees ({BIT}*): Sampling-Based Optimal Planning via the Heuristically Guided Search of Implicit Random Geometric Graphs},
  booktitle = {IEEE International Conference on Robotics and Automation (ICRA)},
  year      = {2015},
  pages     = {3067--3074}
}

@inproceedings{Carvalho2026spam,
  title={Towards Efficient Large-Scale Path Planning for Outdoor Ground Robots},
  author={Carvalho, Afonso E and Portugal, David and Peixoto, Paulo},
  booktitle={2026 IEEE International Conference on Autonomous Robot Systems and Competitions (ICARSC)},
  pages={85--90},
  year={2026},
  organization={IEEE}
}
\end{document}